\documentclass[letterpaper, 10 pt, conference]{ieeeconf}  
\IEEEoverridecommandlockouts

\usepackage{cite}
\usepackage{amsmath,amssymb,amsfonts}
\usepackage{algorithmic}
\usepackage{xcolor}
\usepackage{diagbox}
\usepackage{caption,subcaption}
\usepackage{pifont} 

\usepackage{graphicx}
\usepackage{times} 
\usepackage[font=small,skip=3pt,tableposition=top]{caption}
\usepackage[bookmarks=true]{hyperref}
\usepackage{siunitx}
\usepackage{threeparttable} 

\usepackage{titlesec}

\definecolor{darkgreen}{RGB}{0,127,0}
\definecolor{darkred}{RGB}{200,0,0}
\newcommand{\xmark}{\ding{55}}

\newcommand{\R}{\mathbb{R}}  

\begin{document}

\title{\LARGE \bf NVRadarNet: Real-Time Radar Obstacle and Free Space Detection for Autonomous Driving}
\author{Alexander Popov, Patrik Gebhardt, Ke Chen, Ryan Oldja, \\ Heeseok Lee, Shane Murray, Ruchi Bhargava, Nikolai Smolyanskiy \\
NVIDIA 
}


\maketitle

\begin{abstract}
Detecting obstacles is crucial for safe and efficient autonomous driving. To this end, we present NVRadarNet, a deep neural network (DNN) that detects dynamic obstacles and drivable free space using automotive RADAR sensors. The network utilizes temporally accumulated data from multiple RADAR sensors to detect dynamic obstacles and compute their orientation in a top-down bird's-eye view (BEV). The network also regresses drivable free space to detect unclassified obstacles. Our DNN is the first of its kind to utilize sparse RADAR signals in order to perform obstacle and free space detection in real time from RADAR data only. The network has been successfully used for perception on our autonomous vehicles in real self-driving scenarios. The network runs faster than real time on an embedded GPU and shows good generalization across geographic regions. \footnote{Video at \url{https://youtu.be/WlwJJMltoJY} .}
\end{abstract}


\section{INTRODUCTION}

The ability to detect dynamic and stationary obstacles (e.g., cars, trucks, pedestrians, bicycles, hazards) is critical for autonomous vehicles. This is particularly important in semi-urban and urban settings characterized by complex scenes with large amounts of occlusion and varieties of shapes.

Previous perception methods rely heavily on utilizing cameras~\cite{liftsplatshoot}~\cite{cameraCenternet}\cite{camera10d} or LiDARs~\cite{mvLidarNet}~\cite{hendy2020fishing}~\cite{fastFurious}~\cite{2018pixor} to detect obstacles. These methods have a number of drawbacks: they are unreliable in cases of heavy occlusions, the sensors may be prohibitively expensive, they can be unreliable in adverse weather conditions~\cite{lidarFog} or at night. Traditional RADAR based obstacle detection methods work well in detecting moving objects that have good reflection properties, but tend to struggle when estimating object dimensions and orientations and can often fail in detecting stationary objects or objects with poor RADAR reflectivity.

In this paper, we present a deep neural network (DNN) that detects moving and stationary obstacles, computes their orientation and size, and detects drivable free space from RADAR data alone. We do this in top-down bird's-eye view (BEV) for highway and urban scenarios while using readily available automotive RADARs. Our method relies on RADAR peak detections alone~\cite{classicalRadar2dFFT}~\cite{classicalRadarDopplerFFT} since automotive RADAR firmware provides only this data. In contrast, other approaches~\cite{azimuthRangeTensor}~\cite{probOrientedRadar}, require expensive Fast Fourier Transformation operations on the raw RADAR data cube cross-sections, which are often not available in commercial automotive sensors. 

Our deep learning approach is able to accurately distinguish between stationary obstacles, such as cars, and stationary background noise. This is important when navigating in a cluttered urban environment. In addition, our approach allows us to regress the dimensions and orientations of these obstacles, which classical methods cannot provide. Our DNN can even detect obstacles with poor reflectivity like pedestrians. Finally, our method provides an occupancy probability map to mark unclassified obstacles and regresses drivable free space.

We have tested our NVRadarNet DNN in real-world autonomous driving on our vehicles running NVIDIA DRIVE AGX's embedded GPU. Our DNN runs \emph{faster than real-time} at \textbf{1.5~ms} end-to-end and provides sufficient time for the planner to react safely. 

Our contributions are as follows:
\begin{itemize}
    \item NVRadarNet: A first of its kind multi-class deep neural network that detects dynamic and stationary obstacles end-to-end without post-processing in a top-down bird's-eye view (BEV) using only peak detections coming from automotive RADARs;
    \item A novel semi-supervised drivable free space detection method using only RADAR peak detections;
    \item A DNN architecture that runs \emph{faster than real-time} at \textbf{1.5~ms} end-to-end on an embedded GPU.
\end{itemize}

\section{PREVIOUS WORK}

\textbf{Obstacle Detection.} Fast and efficient obstacle perception is a core component of a self-driving vehicle. Automotive RADAR sensors provide a cost-efficient way of obtaining rich 3D positioning and velocity information and are widely available on most modern cars. Several recent papers examined the use of the dense RADAR data cubes in order to perform obstacle detection~\cite{azimuthRangeTensor}~\cite{probOrientedRadar}. However, these methods require high input/output bandwidth to obtain such rich data, making them difficult to implement in real-world autonomous vehicles. Thus, most classical methods in automotive RADAR applications utilize post-processed peak detections from the data cube in order to perform classification and occupancy grid detection~\cite{radarClassicalBerha}~\cite{radarClassicalEnsamble}~\cite{occupancySemanticGridBMW}. Others realized that the RADAR peak detections can be viewed as a sparse 3D point cloud and therefore can be used in sensor fusion along with 3D LiDAR points in approaches similar to LiDAR DNNs~\cite{mvLidarNet}~\cite{4DNet}~\cite{radarNet2020}~\cite{hendy2020fishing}~\cite{liraNet}. There were also attempts to enhance camera 3D obstacle detection by fusing it with RADAR ~\cite{centerNetCameraRadarFusion}.

\textbf{Free Space Detection.} RADAR-based drivable free space estimation has been attempted in~\cite{ism} and~\cite{occupancyISM2019}. 

Our DNN performs multi-class detection of dynamic and static obstacles together with the segmentation of drivable free space by using RADAR peak detections alone. Our DNN architecture is lightweight and runs \emph{faster than real-time} at \textbf{1.5~ms} end-to-end on an embedded GPU (on NVIDIA DRIVE AGX). It has been proven to be robust in real-world driving and was tested on over $10000$~km of highway and urban roads as part of our autonomous stack. To date, we are not aware of any RADAR peak detections only DNN that can perform all of these tasks and can run efficiently on autonomous vehicles.

\section{METHOD}

\subsection{Input Generation}

The input to our network is a top-down BEV orthographic projection of accumulated RADAR detection peaks around our ego-vehicle, which is placed at the center of this top-down bird's-eye view (BEV) with its front facing right.

To compute this input, we first accumulate RADAR peak detections across all RADAR sensors on our vehicle (8 radars covering 360 degrees field of view, as found in our test fleet) and then transform them to our ego-vehicle rig coordinate system. We also accumulate these peak detections temporally over $0.5$~seconds in order to increase the density of the signal. Each data point gets a relative timestamp to indicate its age, similar to~\cite{4DNet}. Next, we perform ego-motion compensation for the accumulated detections to the latest known vehicle position. We propagate the older points using the known ego-motion of our vehicle to estimate where they will be at the time of the DNN inference (current time).

Next, we project each accumulated detection to a top-down BEV grid using the desired space quantization to create an input tensor for our DNN. We set our input resolution to 800$\times$800 pixels with $\pm$ $100$~m range in each direction, resulting in $25$~cm per pixel resolution. Each valid BEV pixel (with data) gets a set of features in its depth channel computed by averaging the raw signal features of the RADAR detections that land in that pixel. Our final input for time $t$ is a tensor $I_t \in \R^{h \times w \times 5}$ where $h=800, w=800$ are height and width of a top-down view. The 5 RADAR features in the depth channel are the averages of: Doppler, elevation angle, RADAR cross section (\textit{RCS}), azimuth angle and the relative detection timestamp. We normalize these values to a $[0,1]$ range for training stability using maximum and minimum values provided by the hardware specifications. The resulting tensor is used as  input to our network.

\subsection{Label Propagation}\label{label_propogation}
\begin{figure}
    \centering
\medskip
\begin{subfigure}{1\linewidth}
 \includegraphics[width=1.0\linewidth]{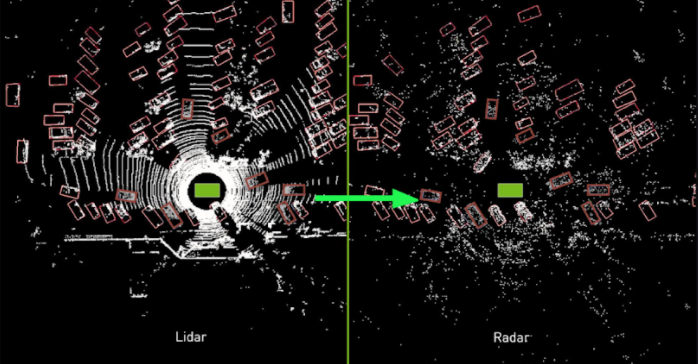}
\end{subfigure}
\caption{Propagating bounding box labels for cars from the LiDAR domain to the RADAR domain.}
\vspace{-2mm}
\label{fig:lidar_to_radar}
\end{figure}

We use LiDAR-based human-annotated bounding box labels as the ground truth for training our RADAR DNN. These labels are created using the LiDAR data for the same scenes on which we train our RADAR DNN.
Given how sparse the RADAR signal is, it is practically impossible for humans to distinguish vehicles using RADAR points alone even in the top-down BEV view. Hence, we rely on LiDAR to label training data. We capture both LiDAR and RADAR data at different frequencies and select the data closest in time for processing. We then create a top-down BEV projection of the LiDAR scene for humans to annotate objects with the bounding box labels and free space with polylines. For each labeled LiDAR BEV frame, we compute the closest accumulated RADAR BEV image via the pre-processing method described above and then transfer the labels to the RADAR top-down view. We further clean up the the ground truth by removing any vehicle labels that contain fewer than $4$ RADAR detections within a $70$~m range. Finally, we remove any detections with an \textit{RCS} below $-40$~dBm as we found that they introduce more noise than signal. An example can be seen in Fig.~\ref{fig:lidar_to_radar}.

\subsection{Free Space Label Generation}

\begin{figure}
    \centering
\begin{subfigure}{1\linewidth}
 \includegraphics[width=1.0\linewidth]{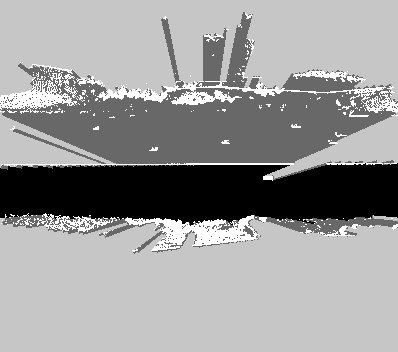}
\end{subfigure}
\caption{Visual representation of the free space target: observed and free in black, observed and occupied in white, unobserved in light gray and partially observed in dark gray.}
\vspace{-4mm}
\label{fig:freespace_labels}
\end{figure}

The free space target is generated by using the raw LiDAR point cloud. First, we pre-process the point cloud by identifying and removing the points belonging to the drivable surface itself by using surface slope angle estimation of adjacent LiDAR scan lines. We then overlay manually obtained LiDAR free space labels to further clean up this estimate. Next, a set of rays is traced from the ego-vehicle's origin in all angular directions, enabling us to reason about which regions are:
\begin{itemize}
  \item Observed and free
  \item Observed and occupied
  \item Unobserved
  \item Partially observed
\end{itemize}

Finally, we overlay our existing 3D obstacle labels on top of the automatically derived occupancy. We explicitly mark obstacles as observed and occupied. See Fig.~\ref{fig:freespace_labels}.

\subsection{Dataset}\label{dataset}

Our model is trained on a diverse internal dataset with over 300k training frames and over 70k validation frames sampled from hundreds of hours of driving in several geographic regions. The dataset includes a combination of urban and highway data and contains synchronized LiDAR, RADAR and IMU readings. The labels are human annotated and include vehicles, cyclists, pedestrians and drivable free space. 

\subsection{Network Architecture}

\begin{figure}
    \centering
\smallskip
\begin{subfigure}{1\linewidth}
 \includegraphics[width=1.0\linewidth]{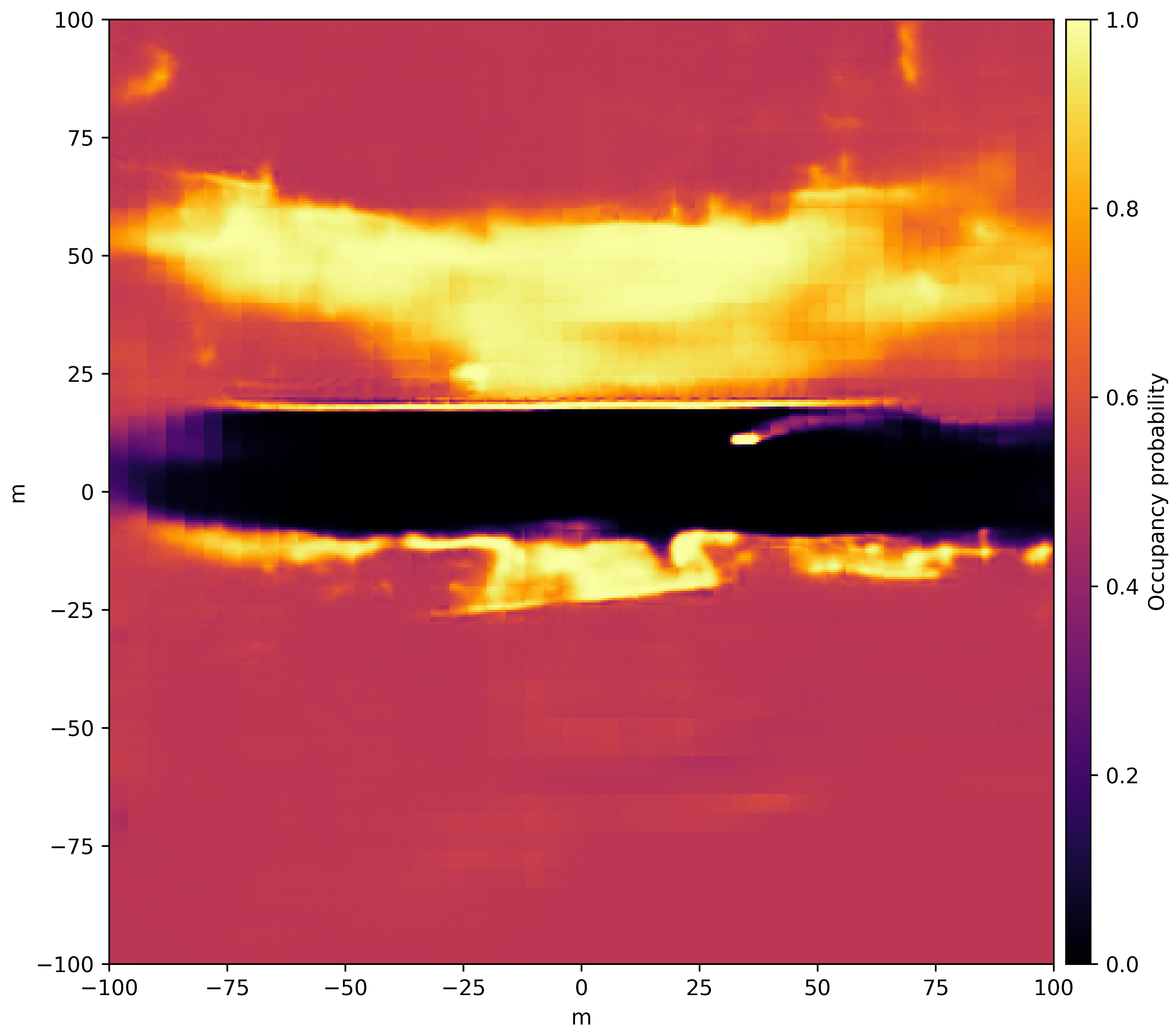}
\end{subfigure}
\caption{Inferred dense occupancy probability map, in Cartesian coordinate space, showing probabilities from low in dark to high probability in gradients from red to yellow (highest).}
\label{fig:occupancy_map}
\end{figure}

\begin{figure*}
    \centering
\medskip
\begin{subfigure}{0.7\linewidth}
 \includegraphics[width=1\linewidth]{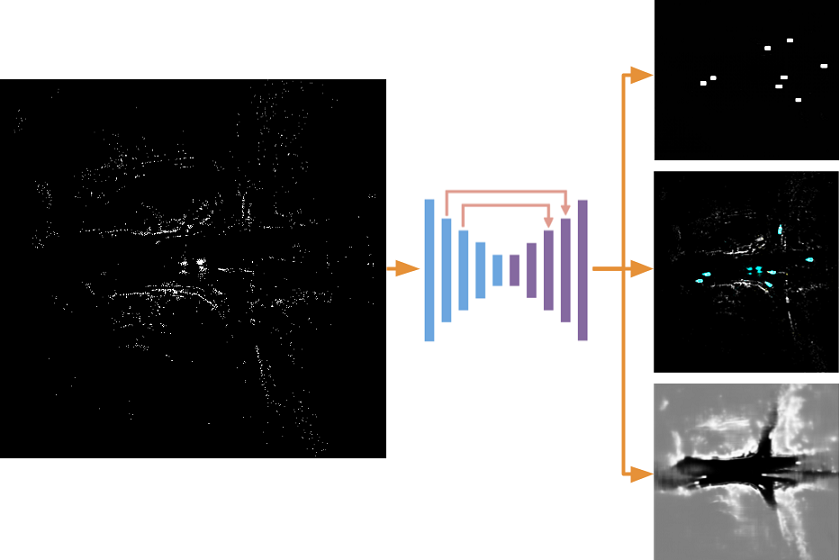}
\end{subfigure}
\caption{Network architecture. Our network uses CNN for the encoder and decoder with skip connections.
The network has three heads: a classification head (produces detection probabilities), a shape regression head (produces bounding box parameters) and a free space segmentation head.}
\label{fig:network_diagram}
\end{figure*}

\begin{table}
\smallskip
\caption{Network architecture for NVRadarNet}
\begin{center}
\begin{minipage}[t]{.95\linewidth}
\resizebox{\linewidth}{!}{
\begin{tabular}{l|l|l|l}
	\textbf{Layer} & \textbf{Layer description} &	\textbf{Input} & \textbf{Output dimensions} \\
\hline 
	\hline
	\multicolumn{3}{c}{\textbf{Inputs:}}	 \\
input & \emph{Input RADAR data}	& -- & $ 5 \times 800 \times 800$  \\
	\hline
	\multicolumn{3}{c}{\textbf{Encoder:}}	 \\
	\hline
1	& conv $(7 \times 7), ReLU $ & input & $ 64 \times 400 \times 400 $ \\
2a	& conv $(3 \times 3), ReLU $ & 1 & $ 64 \times 200 \times 200  $ \\
2b	& conv $(3 \times 3), ReLU $ & 2a & $ 64 \times 200 \times 200  $ \\
3a	& conv $(3 \times 3), ReLU $ & 2b & $ 64 \times 200 \times 200  $ \\
3b	& conv $(3 \times 3), ReLU $ & 3a & $ 64 \times 200 \times 200  $ \\
4a	& conv $(3 \times 3), ReLU $ & 3b & $ 128 \times 100 \times 100  $ \\
4b	& conv $(3 \times 3), ReLU $ & 4a & $ 128 \times 100 \times 100  $ \\
4c	& conv $(3 \times 3), ReLU $ & 4b & $ 128 \times 100 \times 100  $ \\
4d	& conv $(3 \times 3), ReLU $ & 4c & $ 128 \times 100 \times 100  $ \\
5a	& conv $(3 \times 3), ReLU $ & 4d & $ 256 \times 50 \times 50  $ \\
5b	& conv $(3 \times 3), ReLU $ & 5a & $ 256 \times 50 \times 50  $ \\
5c	& conv $(3 \times 3), ReLU $ & 5b & $ 256 \times 50 \times 50  $ \\
5d	& conv $(3 \times 3), ReLU $ & 5c & $ 256 \times 50 \times 50  $ \\
6a	& conv $(3 \times 3), ReLU $ & 5d & $ 512 \times 50 \times 50  $ \\
6b	& conv $(3 \times 3), ReLU $ & 6a & $ 512 \times 50 \times 50  $ \\
6c	& conv $(3 \times 3), ReLU $ & 6b & $ 512 \times 50 \times 50  $ \\
6d	& conv $(3 \times 3), ReLU $ & 6c & $ 512 \times 50 \times 50  $ \\
	\hline
	\multicolumn{3}{c}{\textbf{Decoder:}}	 \\
	\hline
freespace\textunderscore output & deconv $(4 \times 4), {ReLU} $ & 6d & $ 2 \times 400 \times 400 $ \\
	\hline
regression\textunderscore output & deconv $(4 \times 4), {ReLU} $ & 6d & $ 6 \times 200 \times 200 $ \\
	\hline
class\textunderscore output & deconv $(4 \times 4), {ReLU} $ & 6d & $ 4 \times 200 \times 200 $
\end{tabular}
}\end{minipage}

\end{center}
\label{tab:network_arch}
\vspace{-4mm}
\end{table}

We use a DNN architecture similar to a Feature Pyramid Network~\cite{fpn}. Our DNN consists of encoder and decoder components and several heads for predicting different outputs. See Fig.~\ref{fig:network_diagram} for high-level structure and  Table~\ref{tab:network_arch} for details.

Our encoder starts with a \textit{2D convolutional layer} with $64$ filters, stride 2 and $7\times7$ kernels. It is followed by $4$ blocks of $4$ layers each, where each block increases the number of filters by two, while dividing the resolution in half. Each layer in the block contains a \textit{2D convolution} with \textit{batch normalization} and \textit{ReLU activation}.

The decoder consists of one \textit{transposed 2D convolution} with stride $4$ and $4\times4$ kernels per head. We also experimented with using two \textit{transposed 2D convolutions} with a skip connection in the middle. The resulting output tensor is at $1/4$ of the spatial resolution of the input.

We use the following heads in our network:
\begin{itemize}
    \item \textbf{Class segmentation head} predicts a multi-channel tensor, one channel per class. Each value contains a confidence indicating that a given pixel belongs to a class corresponding to its channel.
    \item \textbf{Instance regression head} predicts oriented bounding boxes for an object using an $n_r$ ($n_r = 6$) channels of information for each predicted pixel. The $n_r$ element vectors contains:
[$\delta_x$, $\delta_y$, $w_0$, $l_0$, $\sin\theta$, $\cos\theta$], where ($\delta_x$, $\delta_y$) points toward the centroid of the corresponding object, $w_0$ $\times$ $l_0$ are the object dimensions, and $\theta$ is the orientation in the top-down BEV.
    \item \textbf{Freespace head} computes a map of occupancy probabilities for each grid cell~\cite{ism}.
\end{itemize}

\subsection{Loss}

Our loss consists of a standard cross-entropy loss for the classification head, with a larger weight emphasis on the minority classes, $L_1$ loss for instance bounding box regression, and the inverse sensor model loss for free space detection~\cite{ism}.

We combine these losses using Bayesian learned weights by modeling each task weight as a homoscedastic task-dependent uncertainty following the method described in \cite{multiTaskLearning}. This approach allows us to efficiently co-train these three diverse tasks without affecting the overall model accuracy. 

The total loss is defined as:
\begin{align}
   \textit{TotalLoss} = \sum_{i=0}^{K-1} L_i w_i + \mu_w 
\end{align}
where $K$ is the number of tasks/heads, $L_i$ is a loss for task $i$, $w_i$ = $e^{-\delta_i}$, $\delta_i$ is a learned log variance parameter per task, and $\mu_w$ is the mean of $w_i$ weights.

\subsection{End-to-end Obstacle Detection}

In order to avoid expensive non-maximum suppression (NMS) or clustering at post-processing (e.g. DBSCAN), we employ an end-to-end approach by classifying a single pixel per obstacle, as inspired by OneNet~\cite{oneNet}. 

First we compute the $L_1$ loss for the regression head and the pixel-wise classification loss for the classification head. Next, for each target obstacle, we select the foreground pixel with the lowest total loss between $(\textit{ClassWeight} * \textit{ClassLossPerPixel}) + \textit{RegressionLossPerPixel}$. This pixel is then selected for the final loss computation while the rest of the foreground pixels are ignored. The losses from the background pixels are then selectively used by utilizing hard negative mining. Finally, we perform batch normalization by dividing the total cross-entropy loss by the number of positive pixels selected during the above process. The regression losses are computed only for the selected positive pixels. 

At inference time we pick all of the candidate pixels above a certain threshold in the classification head, per class. The obstacle dimensions are picked directly from the regression head for each corresponding threshold candidate.

By using this technique our network is able to directly output the final obstacle without expensive post-processing.

\subsection{Converting ISM Head Output to a Radial Distance Map}
\begin{figure}
\centering
\smallskip
\begin{subfigure}{1\linewidth}
 \includegraphics[width=1.0\linewidth]{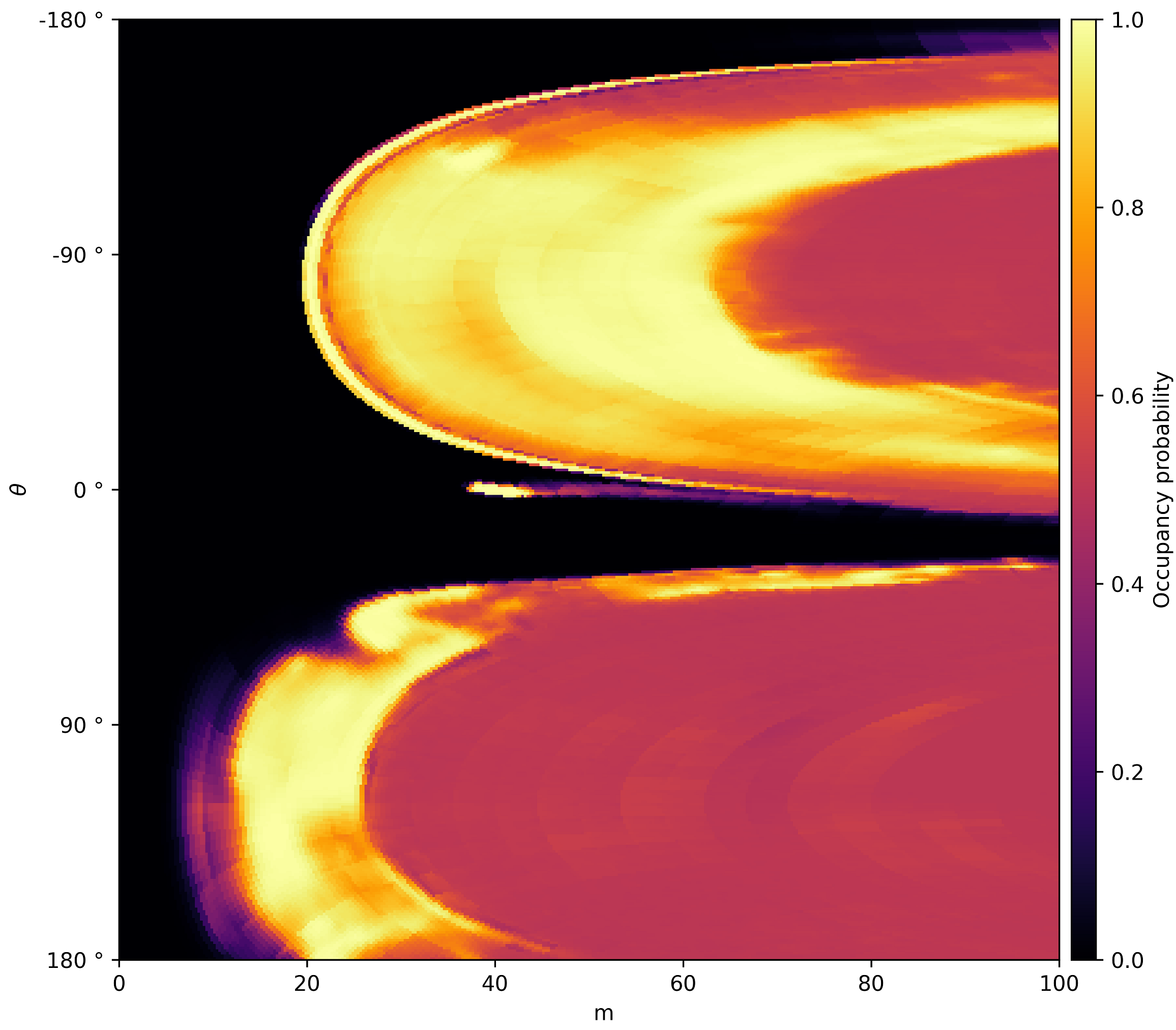}
\end{subfigure}
\caption{Predicted dense occupancy map re-sampled into the polar coordinate system centered around the reference point $\vec{p}_{ref}$. Gradient colors from red (low) to yellow (high) show probabilities.}
\label{fig:occupancy_map_polar}
\end{figure}

\begin{figure}
\vspace{-4mm}
\centering
\begin{subfigure}{1\linewidth}
 \includegraphics[width=1.0\linewidth]{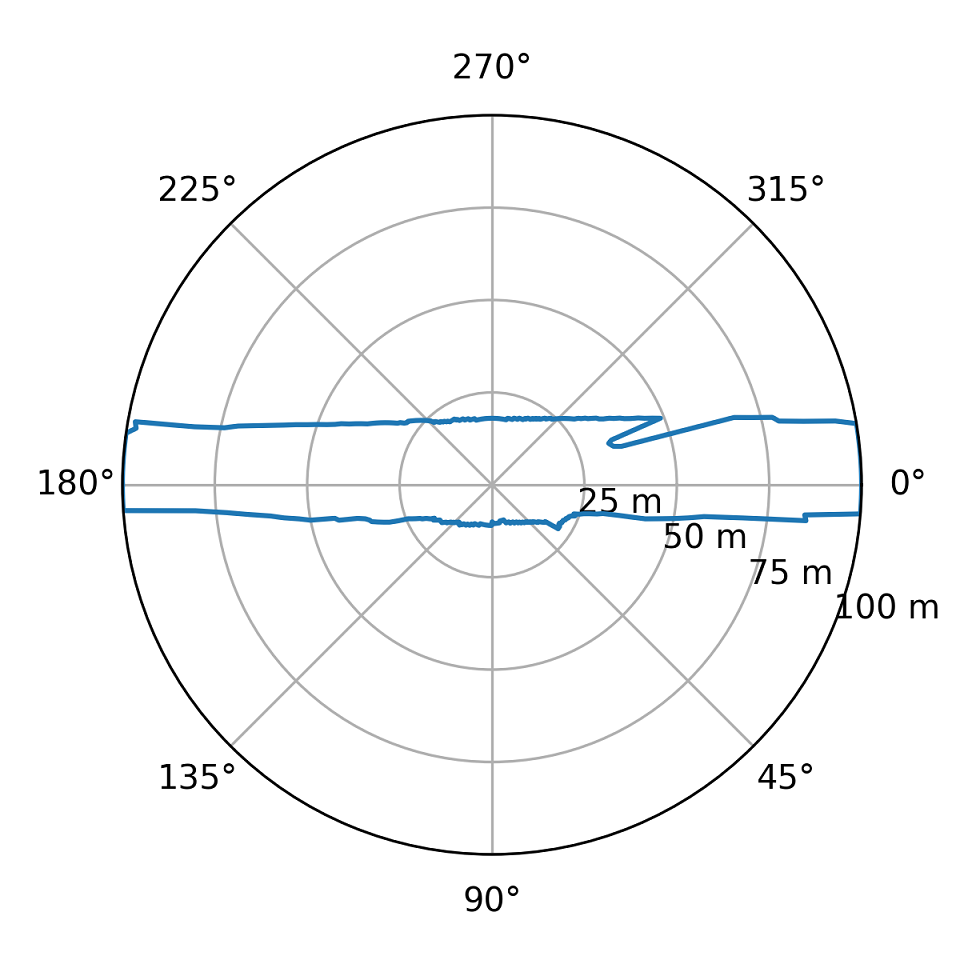}
\end{subfigure}
\caption{Radial distance map representation of the drivable free space boundary extracted from the predicted dense occupancy probability map in Cartesian coordinate space.}
\label{fig:drivable_freespace_boundary}
\vspace{-2mm}
\end{figure}

Autonomous vehicle applications often represent drivable free space area by its boundary contour. In this sections we describe how to convert the boundary contour to a \textit{radial distance map} (RDM) if needed. The RDM assigns a set of angular directions $\phi_{f}$, in the top-down BEV view around the ego-vehicle, to the distance $d_{f}$ between a reference point $\vec{p}_{ref}$ on the ego-vehicle and the drivable free space boundary.  
To compute the RDM, we first re-sample the dense occupancy probability map (DNN output) into a polar coordinate system centered around the reference point $\vec{p}_{ref}$. By employing a nearest-neighbour interpolation schema, the re-sampling process can be expressed in terms of an indexing operation. This assigns the value of each pixel $(\phi_{f},d_{f})$ of the polar representation the value of a single pixel of the predicted dense occupancy probability map. Since this mapping only depends on the dimensions of the occupancy map and the position of reference point $\vec{p}_{ref}$, all required indices can be calculated offline and stored in a lookup table. Fig.~\ref{fig:occupancy_map_polar} shows the occupancy probability map. Fig.~\ref{fig:occupancy_map} shows it re-sampled in polar coordinates. After re-sampling, the distance $d_{f}$ for each angular direction $\phi_{f}$ is determined by finding the first pixel along each angular axis, where the occupancy probability reaches some threshold $p_{occ}$. Fig.~\ref{fig:drivable_freespace_boundary} shows the RDM representation of the drivable free space boundary derived by this procedure from the dense occupancy probability map shown in Fig.~\ref{fig:occupancy_map}.

\section{EXPERIMENTS}

To evaluate our method we trained and tested our model on both the open nuScenes~\cite{nuscenes2019} and on our internally collected datasets.

\subsection{Internal Dataset Experiments}

Datasets, benchmarks and published DNNs dedicated to RADAR based obstacle and freespace detection are limited at this time which presents difficulty when evaluating. See Table~\ref{tab:radarmethods} for a list of available methods and their features. The closest works ~\cite{radarNet2020}~\cite{hendy2020fishing}~\cite{centerNetCameraRadarFusion}~\cite{liraNet} use sensor fusion and do not share RADAR only results publicly. Thus, to the best of our knowledge we are setting a baseline for obstacle detection, classification and freespace regression using RADAR peaks alone. Detection of pedestrians and cyclists is a big challenge due to the sparsity of the RADAR signal.

\begin{table}
\smallskip
\caption{Related RADAR detection methods. Our method (in bold) uses only RADAR data, supports object and free space detection and we provide public results.}
\begin{center}
\begin{tabular}{l|p{0.7cm}|p{0.7cm}|p{1cm}|p{1cm}}
\textbf{Method}&\textbf{Radar Only}&\textbf{Public}&\textbf{Objects}&\textbf{Freespace}\\
\hline
	\hline
RadarNet\cite{radarNet2020} & \xmark & \checkmark & \checkmark & \xmark \\
FishingNet\cite{hendy2020fishing} & \xmark & \checkmark & \checkmark & \xmark \\
LiRaNet\cite{liraNet} & \xmark & \checkmark & \checkmark & \xmark \\
RADModel\cite{azimuthRangeTensor} & \checkmark & \xmark & \checkmark & \xmark \\
XSense\cite{probOrientedRadar} & \checkmark & \xmark & \checkmark & \xmark \\
OccupancyNet\cite{occupancyISM2019} & \checkmark & \checkmark & \xmark & \checkmark \\
\textbf{Ours} & \textbf{\checkmark} & \textbf{\checkmark} & \textbf{\checkmark} & \textbf{\checkmark} \\
\end{tabular}
\end{center}
\label{tab:radarmethods}
\vspace{-4mm}
\end{table}

We evaluated our DNN on our internal NVIDIA's RADAR dataset as well as on the nuScenes public dataset. We also compared our DNN to other published works as much as was practically possible and list all results in this section.

For our internal NVIDIA's RADAR dataset, we used held out test data mentioned in section~\ref{dataset} for evaluation. It is important to note that, even after filtering out the ground truth bounding boxes, which contain too few RADAR peak detections (as described in section~\ref{label_propogation}), we still end up with noisy ground truth labels. For example, there are many instances where vehicles are occluded by other vehicles and so human labelers are not able to create good ground truth from LiDAR data alone. In such cases, RADAR still produces valid returns and some obstacles are correctly classified by our DNN, but are marked as false positives at evaluation. This lowers our precision. Also, LiDAR sensor is mounted higher on the vehicle (roof) compared to the RADAR sensors (bumpers) and therefore some obstacles may be visible by LiDAR but have limited RADAR visibility which leads to noisy RADAR data and labels. This results in false negatives and lowers our recall. 

Our results for the object detection task can be seen in Tables~\ref{tab:internal_results_800},~\ref{tab:internal_mAP}. The metrics for the free space detection task (Table~\ref{tab:internal_freespace}) are calculated for the free space region and the free space RDM independently. The free space region is defined by an occupancy probability $p_{o} < 0.4$.

\begin{table}
\begin{threeparttable}[b]
\medskip
\caption{Our DNN's obstacle detection accuracy on the internal NVIDIA dataset by class and range.} 
\begin{center}
\begin{tabular}{l|l|p{1.2cm}|p{1.2cm}}
	\textbf{Class} &	\textbf{Range} & \textbf{F-score, small res.}\tnote{1} & \textbf{F-score, large res.}\tnote{2} \\
\hline 
\hline
vehicles & 0 --  10  m   & 0.728 & 0.770 \\
vehicles & 10 -- 25  m  & 0.608 & 0.628 \\
vehicles & 25 -- 40  m  & 0.728 & 0.485 \\
vehicles & 40 -- 70  m  & 0.327 & 0.319 \\
vehicles & 70 -- 100 m & 0.225 & 0.216 \\
\hline
pedestrians & 0 --  10  m   & 0.197 & 0.248 \\
pedestrians & 10 -- 25  m  &  0.204 & 0.24 \\
pedestrians & 25 -- 40  m  & 0.145 & 0.174 \\
pedestrians & 40 -- 70  m  & 0.084 & 0.113 \\
pedestrians & 70 -- 100 m & 0.040 & 0.062 \\
\hline
cyclists & 0 --  10  m   & 0.145 & 0.264 \\
cyclists & 10 -- 25  m  & 0.125 & 0.257 \\
cyclists & 25 -- 40  m  & 0.085 & 0.192 \\
cyclists & 40 -- 70  m  & 0.064 & 0.137 \\
cyclists & 70 -- 100 m &  0.065 & 0.114 \\
\end{tabular}
\begin{tablenotes}
\item [1] Input resolution: $800 \times 800 \times 5$.
\item [2] Input resolution: $1024 \times 1024 \times 5$.
\end{tablenotes}

\end{center}
\label{tab:internal_results_800}
\end{threeparttable}
\end{table}

\begin{table}
\medskip
\begin{threeparttable}[b]
\caption{Our DNN's obstacle detection accuracy on the internal NVIDIA dataset by class.}
\begin{center}
\begin{tabular}{l|l|l}
	\textbf{Class} & \textbf{mAP, small resolution}\tnote{1} & \textbf{mAP, large resolution}\tnote{2} \\
    \hline 
	\hline
vehicles & 0.438  & 0.473 \\
pedestrians & 0.039  & 0.057 \\
cyclists & 0.032  & 0.066 \\
\end{tabular}
\begin{tablenotes}
\item [1] Input resolution: $800 \times 800 \times 5$.
\item [2] Input resolution: $1024 \times 1024 \times 5$.
\end{tablenotes}
\end{center}
\label{tab:internal_mAP}
\end{threeparttable}
\end{table}

\begin{table}
\caption{Our DNN's free space regression accuracy on the internal NVIDIA dataset.} 
\begin{center}
\begin{tabular}{l|l|l|l|l}
	\textbf{Resolution} & \textbf{Accuracy} & \textbf{IoU} & \textbf{RDM MAE} & \textbf{RDM IoU} \\
\hline 
	\hline
$800 \times 800 \times 5$ & 0.970 & 0.597 & $3.129$~m & 0.630  \\
$1024 \times 1024 \times 5$ & 0.968 & 0.576 & $2.674$~m & 0.712  \\
\end{tabular}

\end{center}
\label{tab:internal_freespace}
\vspace{-2mm}
\end{table}

\subsection{NuScenes Dataset Performance}

We further train and evaluate our approach on the public nuScenes dataset~\cite{nuscenes2019}. This dataset
contains sensor data from 1 LiDAR and 5 RADARs. However, the sensors in this dataset are from the older generation making direct comparison difficult. The LiDAR sensor used for the nuScenes data collection contains only 32 beams vs 128 beams in our internal dataset. The extra sparsity in this dataset degrades the quality of our auto-generated free space targets. Similarly, the Continental ARS 408-21 RADARs used in nuScenes dataset produce significantly sparser detections when compared to the newer generation Continental ARS430 RADAR sensors we use in our internal dataset. Nonetheless, we demonstrate respectable results, especially at close ranges. See Tables~\ref{tab:nuscenes_results_800},~\ref{tab:nuscnes_map} and~\ref{tab:nuscenes_freespace} for details.

\begin{table}
\smallskip
\begin{threeparttable}[b]
\caption{Our DNN's obstacle detection accuracy on nuScenes dataset by class and range.} 
\begin{center}
\begin{tabular}{l|l|l|l}
	\textbf{Class} & \textbf{Range} & \textbf{F-score, small}\tnote{1} & \textbf{F-score, large}\tnote{2} \\
\hline 
	\hline
vehicles & 0  -- 10  m  & 0.520 & 0.563 \\
vehicles & 10 -- 25  m  & 0.500 & 0.538 \\
vehicles & 25 -- 40  m  & 0.352 & 0.386 \\
vehicles & 40 -- 70  m  & 0.180 & 0.199 \\
vehicles & 70 -- 100 m  & 0.080 & 0.086 \\
	\hline
pedestrians & 0  -- 10  m  & 0.056 & 0.059 \\
pedestrians & 10 -- 25  m  &  0.046& 0.052 \\
pedestrians & 25 -- 40  m  & 0.030 & 0.040 \\
pedestrians & 40 -- 70  m  & 0.016 & 0.024 \\
pedestrians & 70 -- 100 m  & 0.000 & 0.005 \\
   \hline
cyclists & 0  -- 10  m  & 0.050 & 0.030 \\
cyclists & 10 -- 25  m  & 0.068 & 0.066 \\
cyclists & 25 -- 40  m  & 0.059 & 0.066 \\
cyclists & 40 -- 70  m  & 0.044 & 0.041 \\
cyclists & 70 -- 100 m  & 0.000 & 0.000 \\
\end{tabular}
\begin{tablenotes}
\item [1] Input resolution: $800 \times 800 \times 5$.
\item [2] Input resolution: $1024 \times 1024 \times 5$.
\end{tablenotes}
\end{center}
\label{tab:nuscenes_results_800}
\end{threeparttable}
\end{table}

\begin{table}
\medskip
\begin{threeparttable}[b]
\caption{Our DNN's obstacle detection accuracy on nuScenes dataset by class.} 
\begin{center}
\begin{tabular}{l|l|l}
	\textbf{Class} & \textbf{mAP, small resolution}\tnote{1} & \textbf{mAP, large resolution}\tnote{2} \\
\hline 
	\hline
vehicles & 0.245 & 0.280  \\
pedestrians & 0.002 & 0.003  \\
cyclists & 0.005 & 0.004  \\
\end{tabular}
\begin{tablenotes}
\item [1] Input resolution: $800 \times 800 \times 5$.
\item [2] Input resolution: $1024 \times 1024 \times 5$.
\end{tablenotes}
\end{center}
\label{tab:nuscnes_map}
\end{threeparttable}
\end{table}

\begin{table}
\caption{Our DNN's free space regression accuracy on nuScenes dataset.} 
\begin{center}
\begin{tabular}{l|l|l|l|l}
	\textbf{Resolution} & \textbf{Accuracy} & \textbf{IoU} & \textbf{RDM MAE} & \textbf{RDM IoU} \\
\hline 
	\hline
$800 \times 800 \times 5$ & 0.896 & 0.351  & $10.621$~m & 0.394  \\
$1024 \times 1024 \times 5$ & 0.881 & 0.353  & $9.584$~m & 0.441  \\
\end{tabular}

\end{center}
\label{tab:nuscenes_freespace}
\vspace{-2mm}
\end{table}

We further compare our NVRadarNet DNN free space detection accuracy against the method published in~\cite{occupancyISM2019}, which also presents results on the nuScenes dataset. However, this method operates on a grid covering the area in front of the ego vehicle up to $86$~m with $10$~m to each side, and unlike our approach it does not regress or classify obstacles. For this comparison we performed the evaluation on the same image region, while converting the predicted occupancy probability to three classes \textit{Occupied}, \textit{Free} and \textit{Unobserved} as follows.
\begin{itemize}
    \item Occupied: $p_{occ} > 0.65$
    \item Free: $p_{occ} < 0.35$
    \item Unobserved: $0.35 <= p_{occ} <= 0.65$
\end{itemize}
The results are given in Table~\ref{tab:occupancy_net_freespace}. Our DNN outperforms the other method for occupied space regression (best results in bold) and performs similarly on other tasks.

\begin{table}
\smallskip
\caption{Comparison to OccupancyNet~\cite{occupancyISM2019} on nuScenes dataset. Best results in bold.} 
\begin{center}
\begin{tabular}{l|l|l|l|l}
	\textbf{Method} & \textbf{Occupied} & \textbf{Free} & \textbf{Unobs.} & \textbf{mIoU} \\
\hline 
	\hline
OccupancyNet~\cite{occupancyISM2019} & 0.108 & \textbf{0.614} & \textbf{0.593} & \textbf{0.439} \\
Ours @ $800 \times 800$ & \textbf{0.237} & 0.564 & 0.436 & 0.412 \\
Ours @ $1024 \times 1024$ & 0.222 & 0.574 & 0.405 & 0.400 \\
\end{tabular}

\end{center}
\label{tab:occupancy_net_freespace}
\end{table}

\subsection{NVRadarNet DNN Inference}

Our NVRadarNet DNN can be trained in mixed precision mode using INT8 quantization without any loss of accuracy. We export the network using NVIDIA TensorRT and time it on NVIDIA DRIVE AGX's embedded GPU used in our autonomous vehicles. Our DNN is able to achieve \textbf{1.5~ms} end-to-end inference with all three heads. We process all of the surround RADARs, perform obstacle detection and free space segmentation much faster than real-time on the embedded GPU. It was difficult to find other RADAR DNNs inference timings in the literature for direct comparison. We only found that \cite{azimuthRangeTensor} is an order of magnitude slower. 

\section{CONCLUSION}
In this work, we presented NVRadarNet DNN, a real-time deep neural network for obstacle and drivable free space detection from raw RADAR data provided by common automotive RADARs. We benchmarked our DNN on both internal NVIDIA dataset and the public nuScenes dataset and provided accuracy results. Our DNN runs faster than real-time at \textbf{1.5~ms} end-to-end inference time on NVIDIA DRIVE AGX's embedded GPU. To date, we are not aware of any other RADAR only networks that can simultaneously perform obstacle detection and free space regression while running faster than real-time on automotive embedded computers.

\section*{Acknowledgment}

We would like to thank Sriya Sarathy, Tilman Wekel, Stan Birchfield for their  contributions. We would like to acknowledge David Nister, Sangmin Oh, Minwoo Park for their support.

\clearpage
\bibliographystyle{IEEEtran}
\bibliography{NVRadarNet}

\end{document}